# Vers la reconnaissance de mini-messages manuscrits

Emmanuel Prochasson[1], Emmanuel Morin[2], Christian Viard-Gaudin[3]

Université de Nantes[4]

**Abstract**

Handwriting is an alternative method for entering texts which composed Short Message Services. However, a whole new language features the texts which are produced. They include for instance abbreviations and other consonantal writing which sprung up for time saving and fashion. We have collected and processed a significant number of such handwritten SMS, and used various strategies to tackle this challenging area of handwriting recognition. We proposed to study more specifically three different phenomena: consonant skeleton, rebus, and phonetic writing. For each of them, we compare the rough results produced by a standard recognition system with those obtained when using a specific language model to take care of them.

## 1. Introduction

Les SMS (Short Message Service) représentent l'une des NFCE (Nouvelles Formes de Communication Écrite) ayant connue le plus fort développement ces dernières années. Cette modalité de communication permet l'échange de messages entre téléphones portables, mais comme son nom l'indique, la taille du message est très restreinte (autour de 140 octets).

Sous l'influence conjointe des contraintes de longueur de messages, et de celles liées aux dispositifs physiques de saisie que sont les mini claviers des téléphones, il apparaît que les textes produits s'écartent de façon notable des formes traditionnelles de communication écrite : l'objectif du scripteur étant la production d'un message compréhensible le plus court possible. Il en résulte différents phénomènes qui seront illustrés dans la section 4 de ce document dont les plus communs concernent l'écriture rébus, les squelettes consonantiques et la phonétisation de l'écriture.

Afin de rendre le système de saisie plus ergonomique, différentes améliorations sont envisageables. L'une d'entre elles concerne la saisie de texte prédictive (How et Kan 2005, Grover et al. 1998). Elle consiste en une désambiguïsation de la séquence de touches frappées parmi les 3 ou 4 caractères partagés par une touche du clavier. Une solution encore plus avancée consiste à reconsidérer l'interface historique de saisie qu'est le mini-clavier pour lui substituer une entrée manuscrite sous la forme d'un stylet, voire d'un véritable stylo digital relié au réseau GSM. Une telle solution est aujourd'hui technologiquement disponible pour capturer et envoyer des messages sous forme graphique (MMS), toutefois il est nécessaire d'y ajouter une fonctionnalité de reconnaissance de l'écriture manuscrite si l'on veut ramener ce message à une simple chaîne de caractères (SMS).

C'est l'objectif des travaux entrepris dans le cadre de cette étude, ils visent à étendre les fonctionnalités standard des algorithmes de reconnaissance de l'écriture manuscrite en ligne pour améliorer le traitement des SMS. L'apport proposé se situe au niveau de l'intégration de modèles de langages. Ceux-ci permettent d'améliorer significativement les résultats de la reconnaissance en permettant la prise en compte du contexte pour guider la reconnaissance au delà des indices extraits sur les formes à reconnaître (Perraud et al. 2005). Nous avons ainsi identifié trois formes particulières qui affectent les productions SMS et proposons pour chacune d'elles des ressources spécifiques qui sont intégrées au moteur de reconnaissance. Nous comparons les résultats ainsi obtenus avec la mise en œuvre des ressources standard sur un corpus de SMS manuscrits (appelés ici Mini Messages Manuscrits – MIMEMA) dont nous présentons les caractéristiques principales dans la section suivante.

## 2. Collecte d'un corpus de MIMEMA

En ce qui concerne le scripteur, le protocole de collecte repose sur l'utilisation d'un stylo numérique spécifique (de type Nokia) qui enregistre les informations relatives au tracé de l'écriture. Grâce à une caméra miniature placée sous la bille, le stylo numérique reconnaît et enregistre sa position à intervalle de temps régulier. Pour se repérer, le stylo numérique doit être couplé avec un papier spécialement pré-imprimé (de type Anoto) qui est doté d'une grille microscopique. L'écriture est donc disponible sous forme d'une séquence de points représentant la trajectoire de l'instrument d'écriture, il s'agit d'une représentation dite « en-ligne » à différencier des représentations « hors-ligne » comme celles obtenues sous la forme d'une image après avoir scanné un document (Crettez J-P. et Lorette G. 1998). Un support de connexion assure le transfert des informations contenues dans le stylo vers un ordinateur.

*Figure 1. Extrait du formulaire de collecte de MIMEMA*

Nous avons d'abord collecté à partir de sites web spécialisés différentes productions « attestées » de SMS. Les principales formes recensées correspondent i) à de l'écriture rébus (p. ex. `2m1=demain`, `kfé=café`), ii) à l'utilisation de squelettes consonantiques (p. ex. `slt=salut`, `tjs=toujours`) ou iii) à la phonétisation de l'écriture (p. ex. `koi=quoi`, `comen=comment`). L'ensemble des formes ainsi collectées est utilisé pour générer différents formulaires papiers. Dans ces derniers, il est proposé au scripteur soit de recopier ou bien d'écrire librement des SMS avec une écriture en style pré-casé ou libre (cf. figure 1). Le fait de disposer de ces deux styles d'écriture permettra de juger de l'effort technique nécessaire pour en assurer une reconnaissance acceptable.

Cette collecte a été réalisée auprès de 150 personnes (principalement des lycéens et des étudiants des domaines littéraires et scientifiques). Les messages ainsi collectés ont été manuellement nettoyés (suppression des échantillons non valides), segmentés (segmentation du signal d'écriture par champs du formulaire et par mots) et annotés (association au message de son label texte). Après ces opérations, nous disposons d'un corpus de MIMEMA composé de 1 221 messages pour un total de 11 600 mots où 38 462 caractères (cf. tableau 1).

|                 | Style d'écriture pré-casé | Style d'écriture libre |
|-----------------|---------------------------|------------------------|
| Contenu imposé  | 177                       | 174                    |
| Contenu libre   | 493                       | 477                    |
| **Total**       | **670**                   | **551**                |

*Tableau 1. Caractéristiques du corpus de MIMEMA collecté*

À ce niveau, il est prudent de noter que l'usage du stylo numérique peut introduire un biais dans notre démarche. En effet, l'utilisation du stylo numérique pour collecter des MIMEMA induit l'usage d'une interface de saisie différente de celle utilisée pour les SMS. Selon les travaux du linguiste francophone Anis (2003) : « les conditions matérielles de la communication modèlent fortement la forme linguistique des messages ». Si cette hypothèse — qui s'oppose à l'idée de « langage SMS » — était vérifiée, notre protocole serait « contestable ». Quoi que ces aspects, qui relèvent du champ de la sémio-linguistique de l'écrit, dépassent le cadre de ce travail, nous nous devons d'y accorder une attention particulière.

## 3. Évaluation avec un système industriel de reconnaissance

De manière à juger de la complexité potentielle de la reconnaissance des MIMEMA, nous commençons par leur appliquer un système industriel de reconnaissance de l'écriture.

### 3.1. Système de reconnaissance

Dans ce travail, nous utilisons le système de reconnaissance de l'écriture manuscrite en ligne *MyScript Builder* développé par la société VisionObjects[5]. Ce système opère sur un flux d'encre numérique et gère simultanément les problèmes de segmentation et de reconnaissance au niveau caractères et mots. Il est fourni sous la forme d'un kit de développement logiciel (SDK), qui permet dans sa version de base de disposer de plusieurs modèles de langage mais aussi de construire ses propres ressources notamment des lexiques et expressions régulières. Le moteur de reconnaissance est fourni avec deux ressources linguistiques (notée LK) :

- *LK-text* : pour du texte « correctement écrit ». Cette ressource comporte un lexique français et un modèle de langage (lequel est basé sur les n-grammes au niveau mots). Avec cette ressource, il est possible de reconnaître des mots hors lexique.

- *LK-free* : pour du texte dont la nature est inconnue. Il y a uniquement un modèle de langage au niveau caractères associé à cette ressource pour faciliter l'identification des mots inconnus.

---

[5] http://www.visionobjects.com/

Il est possible d'associer au moteur de reconnaissance aucune ou plusieurs ressources. La séquence des mots fournie par le moteur de reconnaissance est disponible sous la forme d'une liste ordonnée de candidats suivant le score de reconnaissance. Dans les expériences que nous présentons, nous ne considèrerons que le premier candidat de la liste. Enfin, précisons que le moteur de reconnaissance est fourni déjà entraîné pour du français de langue générale. Il ignore tout des spécificités de nos messages ou des caractéristiques de nos scripteurs.

3.2. Estimation des performances

Les messages du corpus étant étiquetés globalement et non caractère par caractère, le taux de reconnaissance le plus facile à calculer serait une mesure globale donnant le pourcentage de messages bien reconnus. Toutefois, une telle mesure ne présente pas une finesse d'analyse suffisante, un message peut être plus ou moins mal reconnu, le rendant néanmoins partiellement compréhensible. Il est donc préférable d'évaluer les performances de reconnaissance au niveau des caractères des messages. Pour cela, le taux de reconnaissance (*TR*) proposé est dérivé de la distance de Levenshtein (*D*) calculée entre les deux chaînes, à savoir la séquence de mots placés en première position par le moteur de reconnaissance et le label correspondant au message à reconnaître : *TR = 100(#label - D) / #label*

La distance *D* calcule les coûts d'édition (insertion, substitution et suppression) nécessaires pour passer d'une chaîne à l'autre. Afin de ne pas pénaliser deux fois un problème de sur-segmentation, qui le plus souvent va conduire à une opération de substitution et d'insertion, le coût d'édition est fixé à 1 pour une suppression ou une substitution et à 0 pour une insertion. Ainsi, dans le cas où tous les caractères d'un message seraient reconnus alors *D=0* et *TR=100 %*. Inversement, si aucun caractère n'est reconnu alors *D=#label* et *TR=0 %*. Cette formule permet ainsi de circonscrire le taux de reconnaissance entre 0 et 100 %. En revanche, elle ne pénalise pas la reconnaissance de messages où des caractères sont insérés. La table 2 est une illustration du calcul de *TR* pour un exemple simple.

| Label | `bjr A 2min` | *#label=8* |
|---|---|---|
| Résultat de reconnaissance | `lojr A Zmuin` | *TR=(8-2)/8=0,75* |

*Table 2. Exemple de reconnaissance erronée avec une sur-segmentation sur le 'b', une substitution sur le '2', et un caractère 'u' inséré*

3.3. Résultats obtenus avec le système de base

La table 3 présente les résultats de reconnaissance pour différentes ressources linguistiques et pour des messages manuscrits cursifs et pré-casés. La première colonne correspond à une « utilisation normale » du système de reconnaissance sans aucune ressource linguistique, puis avec la ressource LK-text, et enfin avec la ressource LK-free. Les résultats de la seconde colonne utilisent un lexique additionnel optimal comportant l'ensemble des mots présents dans les messages. Ils définissent une frontière haute pour le système de reconnaissance.

|  | Sans ressource additionnelle (cursive/pré-casé) | Avec le lexique optimal (cursive /pré-casé) |
|---|---|---|
| Aucune ressource | 87 % / 94 % | **96 % / 96 %** |
| Ressource LK-text | 84 % / 90 % | 95 % / 96 % |
| Ressource LK-free | **88 % / 95 %** | 88 % / 95 % |

*Table 3. Taux de reconnaissance au niveau caractères avec le système de base*

Nous observons ici que la reconnaissance des caractères pré-casés conduit à des meilleures performances (90 à 95 % de reconnaissance suivant les ressources linguistiques) que la reconnaissance du texte sans contrainte (84 à 87 %). Ce phénomène n'est bien sûr pas surprenant, la difficulté de segmentation étant bien supérieure dans le cas de textes cursifs. Par ailleurs, l'utilisation de ressources standard permet au mieux d'augmenter très légèrement la reconnaissance (passage de 94 à 95 % avec LK-free), ou bien dégrade significativement les performances (passage de 94 à 90 % avec LK-text). Cela confirme l'importance d'associer des ressources appropriées pour reconnaître ces messages, et qu'en particulier la ressource LK-text n'est pas vraiment adaptée pour cela. L'adjonction du lexique optimal permet (cf. résultats de la seconde colonne) d'améliorer les taux de reconnaissance, et la seule utilisation de ce lexique donne les meilleurs résultats aussi bien sur le pré-casé que libre (96 % / 96 %).

## 4. Modélisation du langage relatif au MIMEMA

Dans cette section, nous commençons par présenter les ressources construites en relation avec les principales formes recensées de MIMEMA. Nous intégrons ensuite ces ressources au système de reconnaissance et les évaluons au regard des résultats obtenus sans connaissance spécifique sur les MIMEMA.

4.1. Squelettes consonantiques

Un squelette consonantique correspond à l'abréviation d'un mot commun charpenté quasi-exclusivement autour de ses consonnes. Dans ce cas, il faut reconstruire le mot pour pouvoir le lire (p. ex. `dvt=devant`). Ce phénomène peut être modélisé à partir des règles de transformations suivantes :

1. Conservation de la première et de la dernière consonne ainsi que des voyelles situées avant la première et après la dernière (p. ex. pour le mot `indépendance` nous conservons `in...ce`).
2. Suppression des voyelles restantes (p. ex. `indpndce`).
3. Suppression des consonnes `l`, `r` et `h` en position faible, c'est-à-dire lorsqu'elles sont situées après une consonne en début de syllabe.
4. Suppression des consonnes `n` et `m` en position faible, c'est-à-dire lorsqu'elles sont situées avant une consonne en fin de syllabe (p. ex. `indpdce`).

Nous avons utilisé ces règles pour construire un lexique de squelettes consonantiques en nous appuyant sur une liste d'adverbes, adjectifs et noms fréquents extraits d'un corpus français de langue générale. Le lexique ainsi obtenu est composé de 3 244 squelettes consonantiques. Afin d'assurer la complétude de ce lexique, nous avons aussi modélisé les règles de transformations sous la forme d'automates stochastiques à états finis. Ces automates ajoutent aux précédentes règles diverses observations que nous avons réalisées sur les SMS à notre disposition (p. ex. il est fréquent de trouver une sous partie du mot non abrégé comme `bjour` pour `bonjour`). D'une manière générale ces automates sont moins restrictifs que la stricte application des règles de transformations.

4.2. Écriture Rébus

Les rébus sont généralement construits par un mélange de lettres et de chiffres. Ici, il faut lire chaque symbole du mot mis en évidence par son nom et ne pas lire le son associé au mot (p. ex. `paC` doit se lire `pa-cé` et non `pa-que`). En raison de la grande créativité possible pour l'écriture rébus, il n'est pas raisonnable de vouloir construire un lexique dédié (Véronis et Guimier de Neff 2006). La modélisation de ce phénomène a donc été réduite à la définition d'automates stochastiques à états finis. Ces automates, plus complexes que pour les squelettes consonantiques, reposent sur la modélisation des observations suivantes :

- Possibilité de mélanger lettres, chiffres et symboles (p. ex. `a+=a plus, 2m1=demain...`) ;
- Forte prédominance de singletons (p. ex. `c=ces, c'est... ; g=j'ai ; 9=neuf...`) ;
- Faible probabilité d'avoir deux chiffres où plus à suivre dans la même forme.

4.3. Phonétisation de l'écriture

La phonétisation de l'écriture est certainement le phénomène le plus complexe à modéliser puisqu'il correspond à des motivations différentes, tout en gardant à l'esprit que la forme produite doit rester compréhensible lorsqu'elle est lue. Il peut s'agir d'une motivation d'abréviation visant à réduire la frappe sur clavier (p. ex. `tro` pour `trop`), d'un simple jeu visant à produire des néographies « amusantes » (p. ex. `bocou` pour `beaucoup`) ou encore d'une simplification ou méconnaissance de l'orthographe, notamment la conjugaison des verbes (p. ex. la transformation des formes `ai, aie, ait, ais, é...` par la simple lettre `é`).

| **Suppression des "e" muets en fin de mots** |
|---|
| Après une voyelle : `/([aeiouyéèàêîâô])(e)$/ → ['{1}']` |
| Après les consonnes, ce n'est pas automatique (p. ex. : françaisE) :<br>`/([bcdfghjklpqrstvwxz][bcdfghjklmnpqrstvwxz])e$/ → ['{1}'] ;`<br>`/([aiou]r)e$/ → ['{1}'] ; /([kmbvl])e$/ → ['{1}']` |
| [voyelle]se → [voyelle]z : `/([aeiouy])se$/ → ['{1}z']` |
| **Retrait des consonnes muettes en fin de mot** |
| `/[tsdp]$/ → ['']` |
| **Transformations en milieu de mot** |
| Supprimer les doubles consonnes :<br>`/ll/ → ["l"] ; /mm/ → ["m"] ; /nn/ → ["n"] ;`<br>`/pp/ → ["p"] ; /rr/ → ["r"] ; /ff/ → ["f"]` |
| Retirer les "h" lorsqu'ils ne sont pas combinés avec c, p ou s : `/([^pcs])h/ → ['{1}']` |
| Remplacer "qu" par "k" : `/qu/ → ["k"]` |
| Remplacer les "c" par "k" lorsqu'ils ne sont pas devant un "e", "h" ou "i" : `/c([^hei])/ → ['k{1}']` |
| "au" → "o" : `/(e) ?au(x) ?/ → ["o"]` |
| "oi" → "oa" : `/oi/ → ["oa"]` |
| [voyelle] s [voyelle] → [voyelle] z [voyelle] : `/([aeiouy])s([aeiouy])/ → ['{1}z{2}']` |
| ai, ais, é, è → é, è : `/ai|é|è|ais$|ait$/ → ["é", "è"]` |
| Retirer les signes diacritiques : `/ç/ → ["c"] ; /î/ → ["i"] ; ...` |
| **Traitement des exceptions** |
| tes, ses, des → té, tè, sé, sè, dé, dè : `/^([dst])es$/ → ['{1}é', '{1}è']` |
| est → é, è : `/^est$/ → ["é","è"]` |

*Table 4. Règles de transformations pour la phonétisation d'un mot*

Face à la complexité du phénomène et pour ne pas introduire du bruit dans le système, nous avons limité sa modélisation à la production d'un lexique. La table 4 présente la liste des règles de transformations pour la phonétisation d'un mot (p. ex. la phonétisation du mot

`cause` produit les formes suivantes : `kause`, `cose`, `kose`, `koz` et `coz`). Ces règles ont ensuite été appliquées sur les 1 202 mots les plus fréquents d'un corpus français de langue générale pour produire un lexique de 3 171 formes.

## 5. Évaluation

Afin d'évaluer l'apport des différentes ressources linguistiques développées, nous avons manuellement classé les messages du corpus MIMEMA correspondant au style d'écriture précasé en quatre catégories : squelettes consonantiques (54 mots/151 caractères), écriture rébus (90 mots/222 caractères), phonétisation de l'écriture (91 mots/327 caractères) et divers.

Les résultats obtenus, présentés dans la colonne centrale de la table 5, avec les ressources proposées dans les sections 4.1 à 4.3 intégrées au système de reconnaissance, sont comparés avec ceux obtenus avec le système doté des ressources standard (colonne de gauche) et des ressources optimales (colonne de droite) comprenant le lexique des formes à reconnaître[6].

|  | Ressources standard | Ressources développées | Ressources optimales |
|---|---|---|---|
| Squelette consonantique | 94,7 % | 98,0 % | 100 % |
| Ecriture rébus | 92,6 % | 92,6 % | 94,6 % |
| Phonétisation de l'écriture | 94,1 % | 94,1 % | 99,3 % |

Table 5. Taux de reconnaissance caractères avec et sans les ressources développées

Nous pouvons constater que les ressources linguistiques apportées au système de reconnaissance bénéficient principalement à la reconnaissance des squelettes consonantiques (environ 38 % des erreurs initiales sont corrigées). Dans le cas de l'écriture rébus et de la phonétisation de l'écriture, l'amélioration n'est pas directement visible. Néanmoins si nous étudions les résultats obtenus en combinant nos ressources avec la seule ressource LK-text, nous pouvons constater l'apport global des ressources apportées au système (cf. table 6).

Alors que les résultats de la table 5 ne présentaient dans le meilleur des cas qu'une légère augmentation de la précision, la différence est ici beaucoup plus nette. La ressource LK-text est peu adaptée à la reconnaissance des phénomènes étudiés, mais cette combinaison indique clairement que les modèles de langage conçus apportent une information significative et adaptée au système de reconnaissance. La combinaison de la ressource LK-text avec nos propres ressources corrige presque 62 % des erreurs initiales pour la phonétisation et 75 % des erreurs initiales pour les rébus.

|  | LK-text | LK-text + ressources développées | LK-text + lexique optimal |
|---|---|---|---|
| Squelette consonantique | 66,2 % | 98,0 % | 98,0 % |
| Ecriture rebus | 69,1 % | 92,1 % | 94,6 % |
| Phonétisation de l'écriture | 75,2 % | 90,5 % | 99,0 % |

Table 6. Taux de reconnaissance utilisant la ressource LK-text

---

[6] Dans cette table, nous n'indiquons que les meilleurs résultats sans préciser les combinaisons de ressources utilisées (c'est-à-dire celles qui sont fournies en standard avec le système de reconnaissance, à savoir : aucune ressource, LK-text ou LK-free) – cf. section 3.1.

## 6. Conclusion

La reconnaissance de l'écriture manuscrite est une branche de la reconnaissance des formes qui pose un tel niveau de difficultés, notamment à cause de la très grande variabilité intra-classe et de la très forte proximité inter-classe des symboles, que la seule perception des formes n'est pas suffisante pour permettre un taux de reconnaissance élevé. Il est fondamental de pouvoir appuyer la reconnaissance sur des connaissances issues de modèles de langage. L'obtention de tels modèles est une tâche qui devient très délicate dès lors que l'on s'intéresse aux nouvelles formes de communication écrite telles que le sont les SMS. À partir de trois phénomènes particuliers affectant la production de SMS manuscrits, nous avons proposé des ressources linguistiques particulières pour les modéliser. Les modèles sont représentés soit par des lexiques spécifiques dont nous expliquons l'obtention, soit par des descriptions sous forme d'automates stochastiques à états finis. Pour l'un de ces phénomènes, les squelettes consonantiques, nous avons pu améliorer de façon très significative les performances du système de reconnaissance en réduisant de 38 % le taux d'erreur au niveau caractères, seul le style pré-casé ayant été étudié dans ce cas. Pour les deux autres, les écritures rébus et la phonétisation de l'écriture, les performances n'ont pas été améliorées. Pour l'écriture rébus, l'automate proposé est sûrement trop générique et ne reflète pas suffisamment précisément la réalité des rébus trouvés dans les SMS. Si nous avions disposé d'une grande base d'apprentissage correspondant aux rébus rencontrés, il aurait été envisageable d'effectuer un apprentissage de ces automates (le corpus réalisé par Fairon C. et Paumier S. (2006) n'était malheureusement pas disponible lors de cette étude). De même, le lexique utilisé pour couvrir la phonétisation s'est appuyé sur un corpus de langue générale où le discours n'est sans doute pas le plus adapté au langage des SMS. Ces remarques offrent des perspectives d'évolutions intéressantes à ce travail qui, à notre connaissance, fait office de pionnier dans ce domaine.

## Références